\documentclass[conference]{IEEEtran}
\IEEEoverridecommandlockouts
\usepackage{cite}
\usepackage{amsmath,amssymb,amsfonts}
\usepackage{booktabs}
\usepackage{algorithmic}
\usepackage{graphicx}
\usepackage{textcomp}
\usepackage{subcaption}
\usepackage{threeparttable}
\usepackage{xcolor}
\def\BibTeX{{\rm B\kern-.05em{\sc i\kern-.025em b}\kern-.08em
    T\kern-.1667em\lower.7ex\hbox{E}\kern-.125emX}}
\begin{document}

\title{Bridging Distribution Learning and Image Clustering in High-dimensional Space
	
\thanks{*Equal Contribution}
}

\author{\IEEEauthorblockN{Guanfang Dong*}
\IEEEauthorblockA{\textit{Multimedia Research Center} \\
\textit{University of Alberta}\\
guanfang@ualberta.ca}
\and
\IEEEauthorblockN{Chenqiu Zhao*}
\IEEEauthorblockA{\textit{Multimedia Research Center} \\
\textit{University of Alberta}\\
chenqiu1@ualberta.ca}
\and
\IEEEauthorblockN{Anup Basu}
\IEEEauthorblockA{\textit{Multimedia Research Center} \\
\textit{University of Alberta}\\
basu@ualberta.ca}
}

\maketitle

\begin{abstract}

Distribution learning focuses on learning the probability density function from a set of data samples. 
In contrast, clustering aims to group similar objects together in an unsupervised manner.
Usually, these two tasks are considered unrelated.
However, the relationship between the two may be indirectly correlated, with Gaussian Mixture Models (GMM) acting as a bridge.
In this paper, we focus on exploring the correlation between distribution learning and clustering, with the motivation to fill the gap between these two fields, utilizing an autoencoder (AE) to encode images into a high-dimensional latent space.
Then, Monte-Carlo Marginalization (MCMarg) and Kullback-Leibler (KL) divergence loss are used to fit the Gaussian components of the GMM and learn the data distribution.
Finally, image clustering is achieved through each Gaussian component of GMM.
Yet, the ``curse of dimensionality'' poses severe challenges for most clustering algorithms.
Compared with the classic Expectation-Maximization (EM) Algorithm, experimental results show that MCMarg and KL divergence can greatly alleviate the difficulty.
Based on the experimental results, we believe distribution learning can exploit the potential of GMM in image clustering within high-dimensional space.

\end{abstract}


\section{Introduction}

The term ``the curse of dimensionality'' is originally introduced by Richard Bellman \cite{hammer1962adaptive}. 
It refers to the computational challenges that arise when optimizing multivariate functions. 
In such cases, the complexity of a brute-force search increases exponentially as the dimensionality grows.
For image clustering, ``the curse of dimensionality'' presents unique challenges, given that images generally reside in a high-dimensional feature space (e.g., a 100$\times$100 pixel grayscale image can be seen as a point in a 10,000-dimensional space).
As dimensionality increases, a set of data points becomes progressively sparse, causing a highly non-uniform distribution.
Thus, many clustering methods that rely on search for optimization encounter various challenges in a high-dimensional space. 
Even if they are not entirely unusable, their accuracy is still compromised.

The aforementioned challenges for image clustering necessitate a rethinking towards the feasibility of search-based clustering algorithms.
Compared with directly searching and optimizing cluster regions, we hope to address this issue indirectly by describing the distribution of images in a high-dimensional space.
Distribution learning aims to identify and describe the distribution of a set of data points.
Specifically, distribution can be described by some statistical models, such as kernel density estimation \cite{rosenblatt1956remarks}, or non-parametric Bayesian methods \cite{seeger2004gaussian}, etc.
However, using such methods make it difficult to relate to clustering problems.
Therefore, in this work, we use Gaussian Mixture Models (GMM) to fit the distribution of images in a high-dimensional space. 
With the sufficient number of Gaussian components, GMM can accurately describe different parts of the data by adjusting the mean and covariance of individual Gaussian distributions.
Also, the number of Gaussian components in GMM can be chosen manually.
From the perspective of clustering, each Gaussian component can be regarded as a cluster. 
We can attempt to obtain the probability density of the data in each Gaussian component to determine which cluster the corresponding data should belong to.

Despite the fact that GMM has the potential to cluster images while learning the distribution, updating the parameters of each Gaussian component in high-dimensional space is a very challenging task.
This is mainly due to the fact that GMM generally employs the Expectation-Maximization (EM) algorithm, iterating between computing the expectation and maximizing to find the local maximum of the log-likelihood.
However, when applied to high-dimensional space, the EM algorithm encounters two primary issues.
First, the EM algorithm is highly sensitive to initial values, and in high-dimensional spaces, it may converge to a local optima. 
Second, the EM algorithm requires updating of the mean and covariance matrices for each Gaussian component during the M-step.
Since this process has polynomial time complexity, the EM algorithm is extremely sensitive depending on the size of the dimension.
These two factors make it difficult for the EM algorithm to converge in a high-dimensional space, in addition it is extremely time-consuming \cite{kingma2013auto}.

To address the abovementioned challenges, we employ Monte-Carlo Marginalization (MCMarg) to update the Gaussian component parameters within GMM. 
MCMarg is a novel algorithm. 
The main idea behind it lies in randomly sampling some unit vectors. 
Then, the sampled vectors are compared with the actual unit vectors, calculating the Kullback-Leibler (KL) divergence, and subsequently correcting the Gaussian component's parameters. 
Since marginalization is utilized, high dimensionality does not lead to a drastic increase in the complexity of the algorithm.

In addition to this, to better observe the latent distribution of images, we first preprocess the images and map them to a latent vector space through an Auto-Encoder (AE). 
Then, after completing the distribution learning, obtaining the probability density for each Gaussian component for the testing data becomes challenging due to the high dimensionality (we used 512 dimensions). 
Therefore, we sample some data from each Gaussian component. 
The k-nearest neighbor method is applied to these sampled data points. 
This approach allows us to indirectly determine the cluster for the test data.
Our experimental results qualitatively and quantitatively demonstrate that the clustering problems can be indirectly addressed by distribution learning. 
At the same time, compared to traditional search-based clustering methods, our approach exhibits superior performance in high-dimensional space. Our main contributions are as follows:

\begin{enumerate}
	\item We investigate the connection between distribution learning and clustering problems, proposing and validating that clustering problems in high-dimensional space can be transformed into a distribution learning problem.
	\item To address the challenge of clustering in high-dimensional space (such as images), we use the MCMarg method and successfully fit the distribution of data in high-dimensional space through GMM.
	\item We introduce a pipeline specifically for clustering in high-dimensional space. Evaluation shows that our method outperforms many famous clustering approaches in high-dimensional space.
\end{enumerate}

\section{Related Work}
\subsection{Classic Clustering Algorithms in High Dimensional Space}
Clustering is a classical problem in data processing. Many algorithms based on statistics and search have been proposed to address this problem.
K-means \cite{macqueen1967some} iteratively assigns samples to the nearest cluster center and updates the cluster centers. 
Agglomerative clustering \cite{sibson1973slink} constructs an agglomerative structure by progressively merging or splitting existing clusters. 
Balanced Iterative Reducing and Clustering using Hierarchies (BIRCH) \cite{zhang1996birch} performs clustering on large datasets by constructing a clustering feature tree to reduce computational complexity.
Spectral clustering \cite{ng2001spectral} utilizes the spectral (eigenvalue) information of data for clustering.
Density-Based Spatial Clustering of Applications with Noise (DBSCAN) \cite{ester1996density} is a density-based clustering algorithm capable of discovering clusters of arbitrary shapes.

However, due to the ``curse of the dimensionality,'' these methods struggle to deal with high-dimensional data for various reasons. 
When data have high dimensionality, K-means becomes sensitive to initial values, and Euclidean distance is ineffective. 
Maintaining the hierarchy in Agglomerative clustering is time-consuming. 
Building the tree structure in BIRCH is highly complex. 
In DBSCAN, the concept of density becomes hard to define. 
Also, Spectral Clustering requires extensive computation.
Although traditional clustering methods still work in a high-dimensional space, their performance is limited. 
Therefore, it is important to find ways to improve clustering performance in a high-dimensional space.
\begin{figure*}[htb]
	\includegraphics[width=\linewidth]{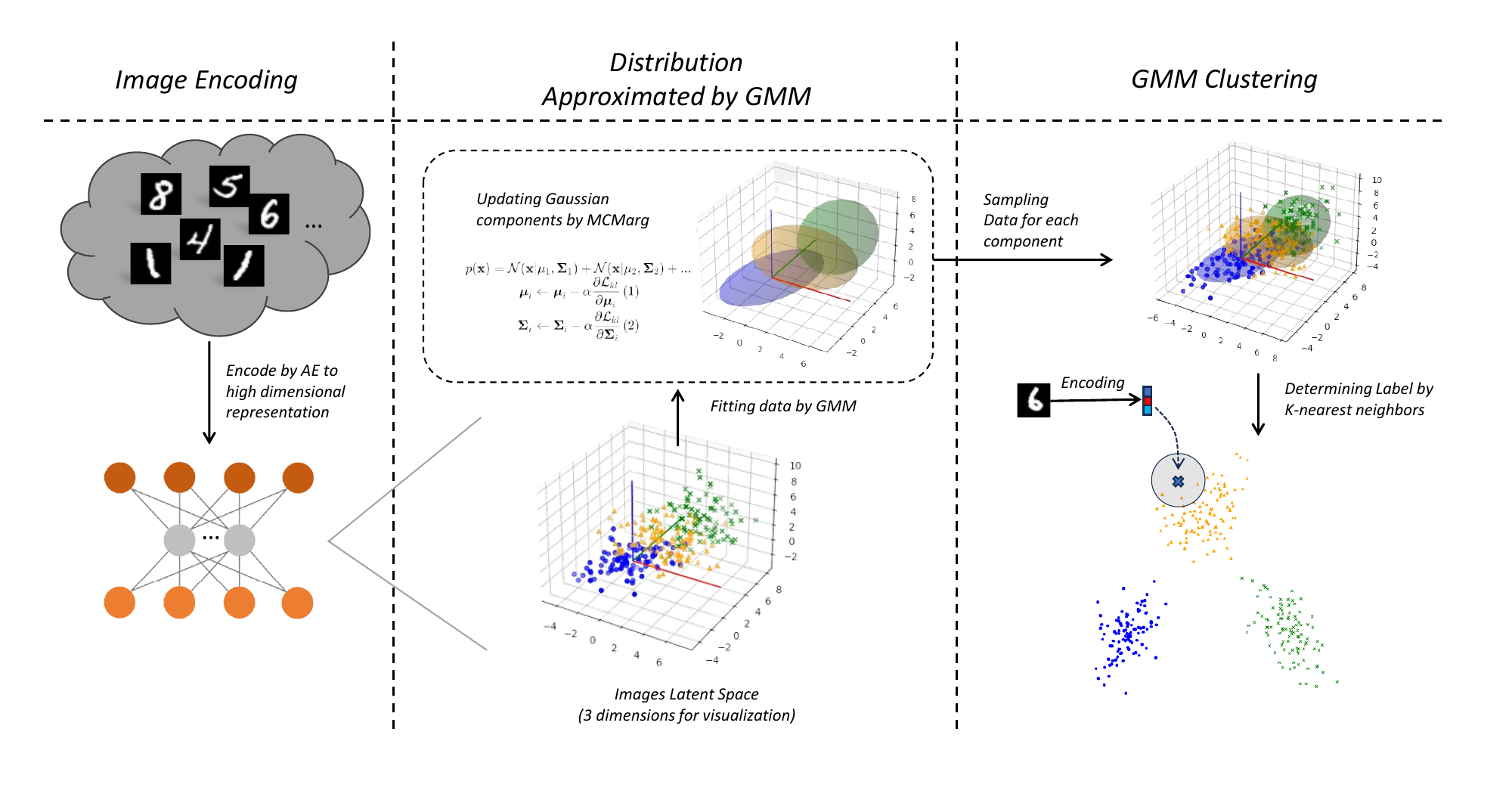}
	\caption{\label{pipeline}
		Workflow of the Proposed Method.}
		\vspace{-10pt}
\end{figure*}
\subsection{Distribution Learning}
Distribution Learning focuses on understanding and modeling the probability distribution of data points. 
After learning the distribution of the data, it can be used for generating new sample points, clustering, classification, and other data analysis tasks.
Although distribution learning seems appealing, there are a limited number of methods for approximating distributions.
Kernel Density Estimation (KDE) \cite{rosenblatt1956remarks} serves as a non-parametric approach to estimate Probability Density Function (PDF).
KDE achieves this by selecting an appropriate kernel function and bandwidth, then computing the density at a given point based on these parameters.
Gaussian Mixture Model (GMM) \cite{bishop2006pattern} is a linear combination of multiple Gaussian distributions, and each component distribution has its own mean, covariance, and mixture weight.
GMM can also be used as soft clustering, where a probability of belonging to each Gaussian component can be calculated.
Recently, a few deep-learning based distribution learning methods have been proposed.
Arithmetic Distribution Neural Network (ADNN) \cite{zhao2022universal} converts distributions to histograms. Then, sum and product distribution layer are used to update histogram distribution kernels.

\subsection{Expectation-Maximization (EM) Algorithm and Monte-Carlo Marginalization (MCMarg) Algorithm}
When discussing GMM, the Expectation-Maximization (EM) \cite{dempster1977maximum} Algorithm is frequently associated.
EM includes two steps, the Expectation step (E-step) and the Maximization step (M-step), iteratively optimizing the parameters of GMM until convergence.
E-step calculates the posterior probabilities of data samples belonging to each Gaussian component. 
M-step updates the weights, means, and covariances of the components by maximizing the likelihood. 
EM algorithm is very efficient to update GMM when data is in low-dimension.
However, as stated in \textit{Section 2.1, Auto-Encoding Variational Bayes} \cite{kingma2013auto}, EM algorithm \textit{``cannot be used''} when a distribution is intractable.
Thus, in this work, we cannot use the EM algorithm to update the GMM parameters.

Monte-Carlo Marginalization (MCMarg) \cite{zhao2023learning} is a newly proposed algorithm to alleviate the converging problem of GMM in high-dimensional space.
MCMarg compares the similarity between the target distribution and GMM distribution in lower-dimensional space from a high-dimensional space through marginalization and random sampling.
Our work is the first to show that the updated strategy of MCMarg can also be applied to clustering problems.

\section{Methodology}
The pipeline of proposed method is illustrated in Figure \ref{pipeline}.
In the initial stage, target images are encoded through an Autoencoder (AE).
Following this, the Gaussian Mixture Model (GMM) is deployed to learn the distribution of latent variables within a high-dimensional space in the AE framework.
During the process of distribution learning, the classic Expectation-Maximization (EM) algorithm is inapplicable for high-dimensional data.
To overcome this limitation, we incorporate the recently developed Monte-Carlo Marginalization (MCMarg) algorithm to update the parameters of each Gaussian component.
From the perspective of clustering problems, each Gaussian component can be regarded as a cluster. 
Thus, due to the difficulty in directly calculating the probability density of a sample point in high-dimensional space, we sampled the points in each Gaussian component. 
By utilizing the k-nearest neighbors method, we can determine to which cluster a test image should belong.
We will introduce details in the following subsections.

\subsection{Problem Statement and Challenge}
The purpose of clustering is to partition data points into $k$ groups or clusters, such that points within the same cluster are similar to each other.
However, in high-dimensional spaces, the complexity of searching for and optimizing cluster regions may increase exponentially with the growth of dimensions.
Thus, in clustering problem, we primarily face two challenges. 
First, as the number of dimensions increases, the observation range in the space expands exponentially, resulting in a sparse data distribution. 
Second, this sparsity leads to the insignificance of differences in distances between data points. 
The distances between data points become difficult to measure accurately using common metrics such as Euclidean distance. 
Therefore, those search-based clustering algorithms perform poorly in high-dimensional space.
Also, based on the aforementioned challenges, we propose to use distribution learning to indirectly solve the clustering task.

\subsection{Image Encoding}
Unlike Deep Embedded Clustering (DEC) \cite{cai2022efficient}, the purpose of using AutoEncoder (AE) in our approach is not to attempt to reduce the dimensionality of images. 
Instead, using AE allow us to unify the measurement standards for images.
Thus, images of different scales can be clustered using the same dimensionality.
In this work, we have employed a latent variable dimensionality of 512. This can be still considered large for a search space.

AE consists of an encoder $f_{\theta}(x)$ and a decoder $g_{\phi}(z)$; the encoder maps an image into a latent vector like $f_{\theta}(x) : \mathbb{R}^{w \times h} \to \mathbb{R}^d$. The decoder performs the opposite operation, mapping from the latent space of dimension $d$ back to the original image space, as $g_{\phi}(z) : \mathbb{R}^d \to \mathbb{R}^{w \times h}$. Suppose we have $n$ input images; by applying the encoding function, we can obtain $n$ points in the latent space, collected into a matrix $\mathbf{Z} = \{\mathbf{z}^{(i)}|i \in [0, n]\} \in \mathbb{R}^{n \times d}$. We will approximate the distribution for $\mathbf{Z}$ in the next step. 

\subsection{Distribution Learning in High-Dimensional Space}
Since our intention to establish an indirect connection between distribution learning and clustering tasks, we employ Gaussian Mixture Model (GMM) to fit the distribution of $\mathbf{Z} \in \mathbb{R}^{n \times d}$ ($d$ is 512 in this work) in the high-dimensional space. 
A GMM represents a mixture of \(K\) multivariate Gaussian distributions. The probability density function is given by:
\begin{equation}
	p(\mathbf{x}) = \sum_{k=1}^{K} \pi_k \mathcal{N}(\mathbf{x}; \mathbf{\mu}_k, \mathbf{\Sigma}_k),
\end{equation}
where \(\mathbf{x}\) is a \(d\)-dimensional vector representing the location of an image in the high-dimensional space. \(K\) is the number of Gaussian components in the mixture. $K$ can also be seen as the number of clusters. 
\(\pi_k\) is the mixing weight of the \(k\)-th Gaussian component. \(\mathcal{N}(\mathbf{x}; \mathbf{\mu}_k, \mathbf{\Sigma}_k)\) is the multivariate Gaussian distribution with mean \(\mathbf{\mu}_k\) and covariance matrix \(\mathbf{\Sigma}_k\).

An optimization method, like the Expectation-Maximization (EM) algorithm or Monte-Carlo Marginalization (MCMarg), can be used to estimate parameters of GMM.
However, the EM algorithm is hard to apply to approximate high-dimensional data for two reasons. 
First, in the M-step, the EM algorithm requires the computation of the covariance matrix for each Gaussian component. In high-dimensional spaces, this matrix can be very large, and the number of parameters to estimate grows quadratically with the increase in dimensions. 
Second, the EM algorithm is highly sensitive to initial parameter values, and in high-dimensional spaces, it is more prone to converge to a local optima rather than the global optimum.
Thus, we apply MCMarg in this work as a first attempt to connect distribution learning and clustering.

Our target distribution from $\mathbf{Z}$ is represented by \( q(\mathbf{z}) \), describing the intractable distribution in high-dimensional space. GMM is shortly represented by \( p(\mathbf{z}; \theta) \), where \( \theta \) is the parameter set. 
MCMarg will select a random unit vector \( \vec{u} \). Then, it will marginalize \( q(\mathbf{z}) \) and \( p(\mathbf{z}; \theta) \) to obtain their marginal distributions \( q_{\vec{u}}(\mathbf{z}) \) and \( p_{\vec{u}}(\mathbf{z}; \theta) \) on the unit vector \( \vec{u} \).
The intention of marginalization is to compare these distributions in a lower-dimensional space.
For the non-differentiable target distribution \( q(\mathbf{z}) \), the samples on the unit vector \( \vec{u} \) are projected, and then the marginal distribution \( q_{\vec{u}}(\mathbf{z}) \) can be estimated by Kernel Density Estimation (KDE).
Finally, the Kullback-Leibler (KL) divergence between the marginal distributions \( q_{\vec{u}}(\mathbf{z}) \) and \( p_{\vec{u}}(\mathbf{z}; \theta) \) can be calculated using the following expression:
\begin{equation}
	\begin{aligned}
		\mathcal{M}(q(\mathbf{z}),p(\mathbf{z};\theta))& =\int_{\vec{u}\in\mathbb{U}}D_{KL}(q_{\vec{u}}(\mathbf{z}\cdot\vec{u})||p_{\vec{u}}(\mathbf{z}\cdot\vec{u};\theta))d\vec{u}  \\
		&\simeq\sum_{\vec{u}\sim\mathbb{U}}D_{KL}(q_{\vec{u}}(\mathbf{z}\cdot\vec{u})||p_{\vec{u}}(\mathbf{z}\cdot\vec{u};\theta)).
	\end{aligned}
\end{equation}
$\mathcal{M}()$ represents the Monte-Carlo Marginalization process. 
$\mathbb{U}$ represents all sampled unit vectors.
The parameters \( \theta \) are optimized using backpropagation algorithms to minimize the KL divergence between the marginal distributions.

\subsection{Utilizing Learned Distributions in Clustering Problems}
For an observed vector \( \mathbf{x} \), we can compute the probability density for each Gaussian distribution since we have obtained the GMM. 
Specifically, we can identify the cluster for \( \mathbf{x} \) by finding the probability density via mean \( \mu_k \) and covariance matrix \( \Sigma_k \) for each Gaussian component.

\begin{equation}
	\begin{aligned}
		\mathcal{N}(\mathbf{x};\mu_k,\Sigma_k)=& \ C\cdot\exp\left(-\frac12(\mathbf{x}-\mu_k)^T\Sigma_k^{-1}(\mathbf{x}-\mu_k)\right) \\
		\text{where} \ C=&\left((2\pi)^{d/2}|\Sigma_k|^{1/2}\right)^{-1}.
	\end{aligned}
\end{equation}
where \(d\) is the dimension of the data, and \(|\Sigma_k|\) is the determinant of the covariance matrix.

However, when $d$ tends to infinity, \( (2\pi)^{d/2} \) will increase exponentially fast. For a finite \( \Sigma_k \), the effect of \( |\Sigma_k| \) will be relatively small. Consequently, the overall value of \( C \) will tend to zero. Mathematically, this can be represented as:
\begin{equation}
	\lim_{{d \to \infty}} C =  \lim_{{d \to \infty}} \left((2\pi)^{d/2}|\Sigma_k|^{1/2}\right)^{-1} = 0
\end{equation}
Although our dimension is not infinity, it can still lead to the issue of precision loss in computer calculations, resulting in a probability density of zero for every Gaussian component ($\left((2\pi)^{512/2}\right)^{-1} = 4.6 \times 10 ^{-205} \approx 0 $).
Thus, we resample the points from each Gaussian component and determine the cluster by nearest-neighbour algorithms. 
For each Gaussian component $k$, we sample a point \(s\) from the corresponding Gaussian distribution \(\mathcal{N}(s;\mathbf{\mu}_k, \mathbf{\Sigma}_k)\). The number of sampled points can be computed as $\left\lfloor N\cdot\pi_k\right\rfloor$, where $N$ is total number of points (60,000 in our case). 
Then, we compute the Euclidean distance between the observation \(\mathbf{x}\) and each sample point, \(d({\mathbf{x}}, {s}_i)\), for \(i=1,\ldots,60,000\). We sort the distances and select the indices of 50 nearest sample points as $
\mathcal{I}_{50} = \{i_1, i_2, \ldots, i_{50}\},$
where \(d({\mathbf{x}}, {s}_{i_1}) \leq d({\mathbf{x}}, {s}_{i_2}) \leq \ldots \leq d({\mathbf{x}}, {s}_{i_{50}})\). For the 50 nearest samples, we compute the vote count for each Gaussian component \(k\) to select the component with the highest vote count as the cluster label for \({\mathbf{x}}\), denoted by $k^* = \arg\max_k \text{Vote Count}_k.$

\section{Experimental Results}
\subsection{Experimental Setting}
Our experiments are conducted on a NVIDIA RTX A4000 GPU. 
Auto-Encoder (AE) and parameters of the Gaussian mixture model are trained independently.
For the Auto-Encoder (AE), we employed the Adam optimizer with a learning rate of 0.001 and applied batch normalization before generating the encoded vector to ensure effective unit vector sampling.
For the Gaussian Mixture Model (GMM), we also used the Adam optimizer with a learning rate of 0.0001, and the number of unit vectors sampled each time was 32.
More implementation details, such as the architecture of the AE, can be found in open-source code\footnote{The code and pre-trained models will be made publicly available on GitHub after the paper is accepted.}.

\subsection{Experimental Design}
\textbf{Dataset}: MNIST and FashionMNIST are used in our experiments. 
MNIST consists of 60,000 training samples and 10,000 test samples, each being a 28$\times$28 pixel grayscale image representing a digit from 0 to 9. FashionMNIST also includes 60,000 training samples and 10,000 test samples, each being a 28$\times$28 pixel grayscale image representing one of 10 different categories of clothing, such as T-shirts, trousers, coats, etc.
We did not preprocess either dataset.
Since clustering is an unsupervised learning task, we only utilized the training subset of the two datasets while disregarding the testing subsets.

\textbf{Baseline Methods}:
In this study, we have chosen K-means  \cite{macqueen1967some}, Agglomerative \cite{sibson1973slink}, Minibatch K-means \cite{mini_batch}, and BIRCH \cite{zhang1996birch} as the baseline methods for comparison. These methods were selected as benchmarks because they are popular unsupervised clustering algorithms widely applicable to various types and scales of datasets. Notably, we did not choose the EM algorithm for GMM parameter updating as a comparison, as stated earlier, since EM cannot effectively perform in high-dimensional spaces.

\textbf{Evaluation Metric}:
The clustering performance of our and selected methods are evaluated by Adjusted Rand Index (ARI).
ARI is a corrected version of the Rand Index (RI) and considers the effect of chance, making it suitable for evaluating the similarity between true and predicted cluster assignments, even with a large number of clusters.
The formula for calculating the Adjusted Rand Index takes into account the combinations of items within the clusters:
\begin{equation}
	ARI = \frac{{\sum\nolimits_{ij} \binom{n_{ij}}{2} - \left[\sum\nolimits_{i} \binom{a_i}{2} \sum\nolimits_{j} \binom{b_j}{2}\right] / \binom{n}{2}}}{{\frac{1}{2} \left[\sum\nolimits_{i} \binom{a_i}{2} + \sum\nolimits_{j} \binom{b_j}{2}\right] - \left[\sum\nolimits_{i} \binom{a_i}{2} \sum\nolimits_{j} \binom{b_j}{2}\right] / \binom{n}{2}}},
\end{equation}
where \( n_{ij} \) is the number of objects in both cluster \( i \) of the true clustering and cluster \( j \) of the predicted clustering. \( a_i \) is the sum of \( n_{ij} \) over all \( j \) for a fixed \( i \). \( b_j \) is the sum of \( n_{ij} \) over all \( i \) for a fixed \( j \).

\subsection{Experimental Results}
\textbf{Quantitative Results}:
In this section, the quantitative evaluation of the proposed approach is discussed. We compare the proposed approach with several classical unsupervised clustering methods.

In order to achieve a good approximation of the intractable distribution, the number of Gaussian components is very important. 
Overall, the selection of the number of Gaussian components in GMM implicates a complex interplay between model complexity, approximation ability, computational efficiency, and generalization capability. 
From the perspective of bias-variance tradeoff, a smaller number of Gaussian components may result in a model that is too simple to capture the complex structure of the data, leading to an increase in model bias. 
Also, a larger number of Gaussian components may make the model too complex, leading to an increase in model variance.
Due to the elevated complexity and diversity inherent in high-dimensional data, employing 64 Gaussian components might be more suitable at capturing the latent structure of the data.

\begin{figure*}[h!]
	\centering
	\begin{subfigure}{0.3\textwidth}
		\includegraphics[width=\textwidth]{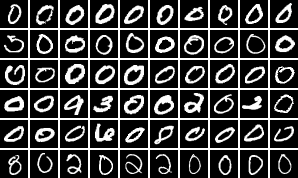}
		\caption{Clusters that contains 0}
	\end{subfigure}
	\hfill
	\begin{subfigure}{0.3\textwidth}
		\includegraphics[width=\textwidth]{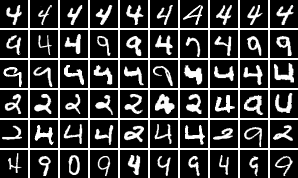}
		\caption{Clusters that contains 4}
	\end{subfigure}
	\hfill
	\begin{subfigure}{0.3\textwidth}
		\includegraphics[width=\textwidth]{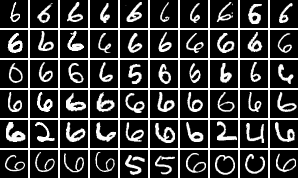}
		\caption{Clusters that contains 6}
	\end{subfigure}
	\caption{MNIST Clusters. Each row of images belongs to the same cluster.}
	\label{fig:mnist_clusters}
	\vspace{-10pt}
\end{figure*}

\begin{figure*}[h!]
	\centering
	\begin{subfigure}{0.3\textwidth}
		\includegraphics[width=\textwidth]{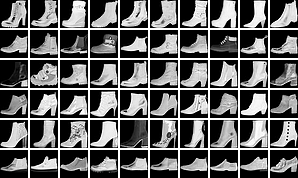}
		\caption{Clusters that contains Ankle Boot}
	\end{subfigure}
	\hfill
	\begin{subfigure}{0.3\textwidth}
		\includegraphics[width=\textwidth]{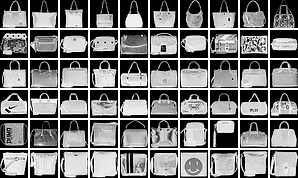}
		\caption{Clusters that contains Bag}
	\end{subfigure}
	\hfill
	\begin{subfigure}{0.3\textwidth}
		\includegraphics[width=\textwidth]{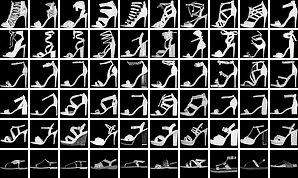}
		\caption{Clusters that contains Sandal}
	\end{subfigure}
	\caption{Fashion MNIST Clusters. Each row of images belongs to the same cluster.}
	\label{fig:fashion_mnist_categories}
	\vspace{-10pt}
\end{figure*}
In addition, for fair comparisons, we also set the same number of clusters for the compared algorithms. The comparison results are shown in Table \ref{tab:results}. As shown in the table, we achieve the best results compared to these famous unsupervised clustering methods. Importantly, the input for both the proposed approach and the compared methods is the hidden latent space derived from an Autoencoder (AE), with a dimensionality of 512. Since the input data is consistent across all methods, the comparison is conducted under fair conditions. The potential of clusters with the view of distribution learning is thus demonstrated, especially considering the high-dimensional input space of 512 dimensions derived from AE.

\begin{table}[ht]
	\centering
	\caption{Comparison of Adjusted Rand Index (ARI) for Different Clustering Methods on MNIST and Fashion MNIST Datasets$^1$.}
	\begin{tabular}{l|cc}
		\toprule
		\textbf{Method}                           & \textbf{Fashion MNIST}        & \textbf{MNIST} \\
		\midrule
		K-Means Clustering               & 0.1634           & 0.1707            \\
		Agglomerative Clustering & 0.1696           & 0.2166   \\
		Mini-Batch K-Means     & 0.1598           & 0.1822            \\
		BIRCH & 0.1742           & 0.2132   \\
		\midrule
		Proposed Method$^2$ & \textbf{0.3606}  & \textbf{0.2491}            \\
		\bottomrule
	\end{tabular}
	\begin{tablenotes}
		\scriptsize
		\item 1. Latent vector is 512-dimension, measured in 64 clusters.
		\item 2. Monte-Carlo Marginalization with Gaussian Mixture Model.
	\end{tablenotes}
	\label{tab:results}
	\vspace{-10pt}
\end{table}

\textbf{Qualitative Results}:
Figures \ref{fig:mnist_clusters} and \ref{fig:fashion_mnist_categories} illustrate the clustering results of our method on different clusters belonging to the same category. When clustering the MNIST and FashionMNIST datasets into 64 clusters, many clusters share the same label but have distinct appearances. However, similar clusters generally exhibit similar appearances. 
For instance, in Figure \ref{fig:mnist_clusters} (a), the last row of `0' is noticeably slimmer, whereas the second-to-last row's `0' is visibly more right-leaning. Similarly, the test results for Fashion MNIST demonstrate a comparable scenario.

Clustering is an unsupervised learning method, with the goal of grouping similar samples together without utilizing any prior label information. 
In contrast, labels are often annotated manually, reflecting specific aspects or categories of the data. 
To some extent, clustering may capture more information than labels. 
However, since clustering relies on the inherent structure of the data, differences can arise. 
For example, in FashionMNIST, long pants and shorts might be similar in shape and texture, thus they may be grouped into the same cluster, even if their labels are different.
Similarly, visual similarity and semantic differences may also lead to such situations. 
Clustering is generally based on visual similarity, like in MNIST, where italic and regular digits might visually differ but carry the same semantic label. 
Clustering might separate them into different groups because they differ visually. 
Meanwhile, the 512-dimensional hidden variable space may capture features not entirely relevant to human annotation, possibly making the distinctions more pronounced.

\textbf{Impact of Sampling Quantity for Points}:
\begin{table}[h]
	\centering
	\caption{The relationship between total sampling points in GMM on FashionMNIST and the ARI index. The quantities represent the total number of samples across all Gaussian components.}
	\begin{tabular}{c|ccccc}
		\toprule
		\textbf{Total Samples} & 1,000 & 5,000 & 10,000 & 60,000 & 600,000 \\
		\midrule
		\textbf{ARI} & 0.2952 & 0.3187 & 0.3453 & \textbf{0.3606} & 0.3426 \\
		\bottomrule
	\end{tabular}
	\label{tab:ARI}
	\vspace{-5pt}

\end{table}

We also investigate at how the number of sampling in GMM affects which cluster an input vector belongs.
Intuitively, more points sampled should yield better results. 
However, we find that the best clustering results occur with 60,000 total samples.
We think this is due to the voting mechanism.
In our implementation, each observation point is matched with its closest 50 samples. 
Its cluster label is then decided based on how many votes each Gaussian component gets.
With the increasing sample points, they will more densely distributed in the space.
Due to the fact that the distribution describes all possible values and their corresponding probabilities for a random variable, having too many samples could introduce more randomness and error.
In high-dimensional spaces, too many samples can weaken the voting impact from Gaussian components with smaller weights.
Gaussian components with larger weights could have too much impact in the vote.
\vspace{-5pt}
\section{Conclusion}
In this work, we study the indirect relationship between distribution learning and clustering.
Although ``the curse of dimensionality'' presents some challenges to image clustering, we indirectly alleviating the issue by fitting the high-dimensional distribution using Gaussian Mixture Models (GMM) and Monte-Carlo Marginalization (MCMarg).
Quantitative results show that our approach achieves competitive results compare with many classic unsupervised clustering methods based on search or hierarchical modeling.
Qualitative results also show that our method can generate meaningful clusters.
Thus, we believe our method could be a potential clustering method in a high-dimensional space.

\bibliographystyle{IEEEtran}  
\bibliography{ref}

\end{document}